\documentclass{article}
\usepackage{spconf,amsmath,graphicx}
\usepackage{amssymb,amsfonts}
\usepackage{multirow}
\usepackage[capitalise]{cleveref}
\usepackage{caption}
\usepackage{subcaption} 
\usepackage{siunitx}
\usepackage{xurl}

\usepackage{flushend}

\title{
ClothFit: Cloth-Human-Attribute Guided Virtual Try-On Network Using 3D Simulated Dataset
}
\def\correspondingauthor{\sthanks{Corresponding author: sungho.suh@dfki.de}}

\name{Yunmin Cho $^{1,2}$, Lala Shakti Swarup Ray $^{1,2}$, Kundan Sai Prabhu Thota $^{2,3}$, Sungho Suh$^{1,2}$\correspondingauthor, Paul Lukowicz$^{1,2}$
\thanks{This work was supported by the BMBF (German Federal Ministry of Education and Research) in the VidGenSense (01IW21003). The Carl-Zeiss Stiftung also funded it under the Sustainable Embedded AI (P2021-02-009).}}
\address{$^{1}$ Department of Computer Science, RPTU Kaiserslautern-Landau, Kaiserslautern, Germany\\
	$^{2}$ German Research Center for Artificial Intelligence (DFKI), Kaiserslautern, Germany\\
        $^{3}$ Shapematchr GmbH, Berlin, Germany }
\begin{document}
%
\maketitle
\begin{abstract}
Online clothing shopping has become increasingly popular, but the high rate of returns due to size and fit issues has remained a major challenge. To address this problem, virtual try-on systems have been developed to provide customers with a more realistic and personalized way to try on clothing. In this paper, we propose a novel virtual try-on method called ClothFit, which can predict the draping shape of a garment on a target body based on the actual size of the garment and human attributes. Unlike existing try-on models, ClothFit considers the actual body proportions of the person and available cloth sizes for clothing virtualization, making it more appropriate for current online apparel outlets. The proposed method utilizes a U-Net-based network architecture that incorporates cloth and human attributes to guide the realistic virtual try-on synthesis. Specifically, we extract features from a cloth image using an auto-encoder and combine them with features from the user's height, weight, and cloth size. The features are concatenated with the features from the U-Net encoder, and the U-Net decoder synthesizes the final virtual try-on image. Our experimental results demonstrate that ClothFit can significantly improve the existing state-of-the-art methods in terms of photo-realistic virtual try-on results. 

\end{abstract}
\begin{keywords}
Virtual try-on, Cloth simulation, Garment fitting, Digital apparel
\end{keywords}
\section{Introduction}
\label{sec:intro}

Image-based virtual try-on has become an increasingly popular method for customers to purchase clothing online, as it allows them to visualize how a particular garment would look on their body before making a purchase. In the e-commerce clothing market, customers often encounter issues with clothing size and fit, which leads to a high rate of returns. According to recent reports \cite{return2sender, 3DLook}, the biggest challenges related to apparel e-commerce non the standardized use of size charts by retailers, which makes it complicated for customers to find the proper fit and different body shapes and individual body proportions makes it challenging to visualize how the clothing would look on the customer. These challenges have led to a growing interest in virtual try-on systems, which provide customers with a more realistic and personalized way to try on clothing. However, accurately simulating the fit of a garment on a customer's unique body shape is still a challenging problem. Existing virtual try-on methods \cite{han2018viton, wang2018toward, neuberger2020image, yang2020towards, fele2022c} have focused on realistic visualization of clothed human images, but have not adequately addressed the issue of garment fit. 

One of the main motivations of our work is to provide a more accurate and realistic virtual try-on experience for users. The ability to simulate the fit of a garment on a user's body can help reduce the number of returns and increase customer satisfaction, ultimately benefiting both customers and online retailers. Recently, many virtual try-on methods have been proposed to utilize deep learning techniques to generate realistic clothing images. For example, Han et al. \cite{han2018viton} proposed a multitask encoder-decoder network, called VITON, to warp the target garment on the target human clothed area without using any 3D information. Similarly, Fele et al. \cite{fele2022c} proposed a context-driven virtual try-on network (C-VTON) to align the target clothing to the segmented body parts with geometric matching. However, many of these methods still suffer from limitations when it comes to accurately simulating the fit of a garment on a user's body. 
Some methods rely on pre-defined garment templates, which may not accurately capture the complexity of real-world garments. Other methods may require extensive user input, such as 3D scans of the user's body, which can be time-consuming and expensive.
Our earlier work \cite{thota2022estimation} proposed an estimation method of 3D body shape and clothing measurements from frontal and side view images to address the size issue.

\begin{figure*}[!t]
    \centering
    \includegraphics[width=1\textwidth]{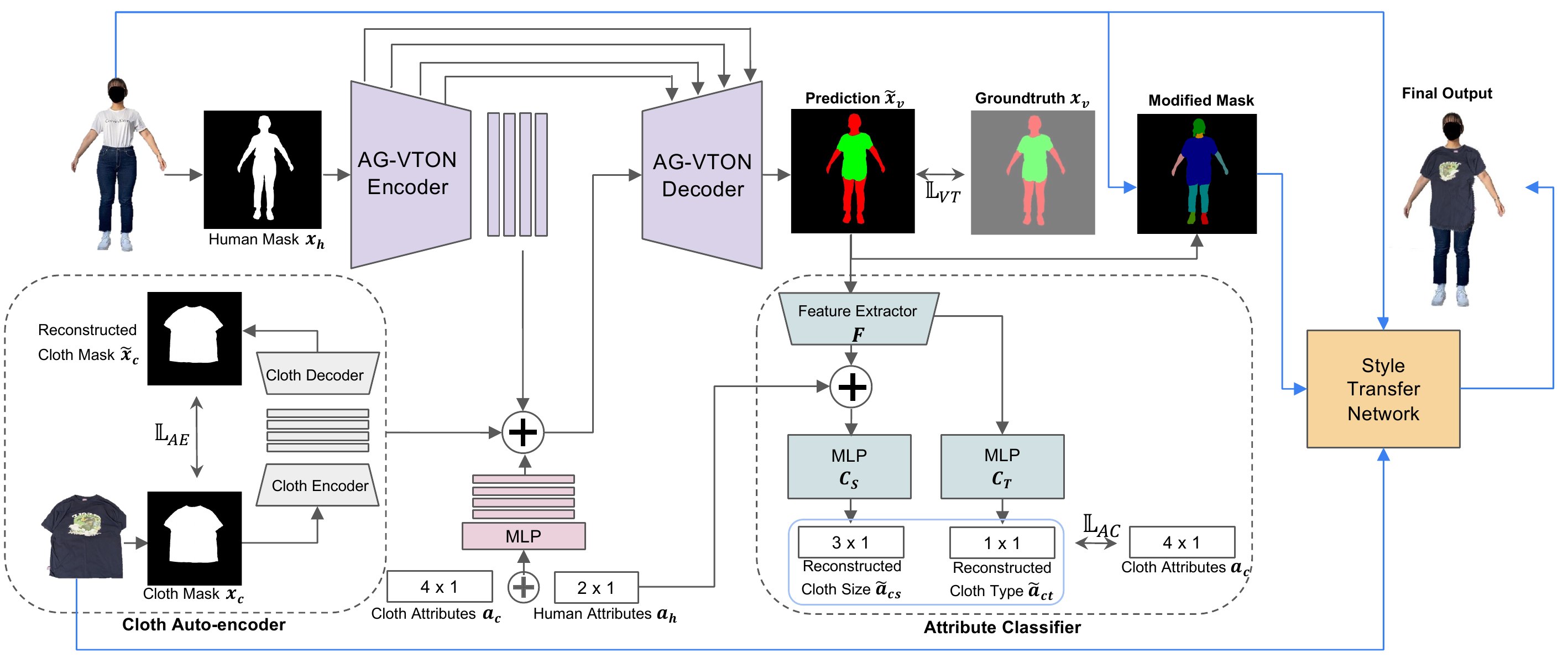}
    \vspace{-7mm}
    \caption{ClothFit architecture: An auto-encoder extracts the features from a cloth mask and an attribute classifier estimates cloth attributes. A U-Net-based attribute-guided virtual try-on network (AG-VTON) generates a masked image of a clothed human using input cloth image, cloth and human attributes, and a frontal-view image of the user. A style transfer network with a modified segmented mask generates the final output image of the user with the input cloth.}    
    \vspace{-5mm}
    \label{fig:clothfit}
\end{figure*}


In this paper, we propose a novel virtual try-on method that utilizes a U-Net-based \cite{ronneberger2015u} network architecture to estimate the fit of a garment on a user's body.
As an extension to body shape estimation from RGB images, this method predicts the draping shape of the garment on the estimated 3D body shape.
The proposed method takes as input a frontal image of the user with height and weight information, as well as an image of the clothing they wish to purchase along with its size factors. We utilize an auto-encoder network to extract the feature vector of the clothing image, and an MLP-based feature extractor to extract the features of the garment attributes and user factors. We then concatenate these features with the features from the U-Net encoder and use the U-Net decoder to synthesize a virtual try-on image. To train our proposed networks, we generate synthetic images simulated over 3,801 people with four different cloth types using the 3D physics simulator Blender \cite{blender} to overcome the limitation of collecting various cloth types, sizes, and human body shapes. Our proposed method is a cloth-human-attribute-guided virtual try-on network (AG-VTON), and we utilize a style transfer network to generate a photo-realistic try-on image. The experimental results demonstrate that the proposed method can synthesize the photo-realistic results by changing the attributes of the user and the target garment, and provides more realistic results than other state-of-the-art methods, which does not show any difference between two different sizes of the cloth.

The main contributions of our study can be summarized as follows: (1) A large synthetic dataset with frontal images of 3801 body types with different height and weight attributes wearing 45612 cloth types along with cloth attributes. (2) A novel virtual try-on method, called ClothFit, to predict the draping shape of the garment over a target body based on the actual size of the garment and human attributes. (3) Validation of the proposed framework with both synthetic and real-world datasets.

The remainder of this paper is organized as follows. \cref{sec:proposedmthod} provides an overview of the network architecture and training process. \cref{sec:experiments} presents our experimental results and analysis. Finally, in \cref{sec:conclusion}, we conclude our work and give an insight into future work.

\vspace{-3mm}
\section{Proposed Method}
\label{sec:proposedmthod}
The proposed framework consists of four networks: an auto-encoder for extracting features of a target clothing image, an attribute classifier for estimating the actual size of the clothing from a clothed human image, a U-Net-based attribute-guided virtual try-on network (AG-VTON), and a style transfer network incorporating the cloth mask and the input user image. The overview of the proposed framework is shown in \cref{fig:clothfit}, where the input is the user image and cloth image of the size of 512 x 512, and the output the clothed human image. 

\subsection{Generating Synthetic Data using 3D Simulation}
To effectively train the proposed cloth-human-attribute-guided virtual try-on networks, we require a large dataset that includes human images, cloth images, and attributes such as human factors and cloth sizes. However, no existing virtual try-on dataset has all of these components. To address this issue, we generate synthetic images using the 3D physics simulator, Blender \cite{blender}. Our dataset comprises three types of images: a target clothing image, a frontal-view image of the user, and a clothed human image. We also include two types of attributes: cloth attributes, including cloth type, chest circumference, total length, and sleeve length, and human attributes, including height and weight of the user. To generate the dataset, we first created 3D SMPL bodies \cite{loper2015smpl} of varying heights and weights using the SMPL-X Blender add-on. Next, we designed cloth sewing patterns for each type of cloth based on their actual design. The sewing patterns were used to join together the different cloth pieces, which were then deformed over the SMPL body mesh to produce the clothed human images \cite{clothaddon}. To facilitate differentiation between the body and cloth, different textures were assigned to the body and cloth. The garment sewing pattern's size was modified to reflect different sizes of the garment. We automated the entire process of rendering cloth and collecting attribute data using Blender API, enabling us to generate a large dataset with a wide range of cloth sizes and body shapes.


\subsection{Attribute-Guided Virtual Try-On}
In the proposed network architecture, we train three networks: a cloth auto-encoder, a U-Net-based AG-VTON, and an attribute classifier. First, the cloth auto-encoder and attribute classifier are trained separately. The cloth auto-encoder takes a cloth mask image of size 512 $\times$ 512 as input and extracts a feature vector of size $\frac{W}{16} \times \frac{H}{16} \times 256$. The encoder comprises four convolution and max-pooling layers, and the decoder contains four upsampling and convolution layers. The loss function of the cloth auto-encoder is defined as follows.
\begin{equation}
    \begin{split}
    \mathbb{L}_{AE} &= \mathop{\mathbb{L}_{BCE}}(\tilde{x}_c, x_c)\\
    \text{where }&\mathop{\mathbb{L}_{BCE}}= \mathop{\mathbb{E}_{p,q}} [q\log p + (1-q)\log (1-p)],
    \end{split}
\end{equation}
where $x_{c}$ and $\tilde{x}_c$ denote the cloth mask image and reconstructed image, respectively, and $\mathop{\mathbb{L}_{BCE}}$ denotes the standard binary cross-entropy loss (BCE).

The attribute classifier is used to classify the type of cloth and estimate the size of the cloth in the clothed human image. At the first stage, the attribute classifier is trained with the ground truth of the clothed human images, and at the second stage, the pretrained attribute classifier is used to improve the performance of the AG-VTON. The attribute classifier takes a clothed human mask image of size $512 \times 512 \times 2$ and the human attributes as input and outputs the cloth attributes as depicted in \cref{fig:clothfit}. The attribute classifier contains a feature extractor from ResNet-18 \cite{he2016deep} and two multilayer perceptrons (MLP) for cloth type classification and cloth size estimation. The loss function of the attribute classifier is expressed as follows:
\begin{equation}
    \begin{split}
        \mathbb{L}_{AC} &= \mathop{\mathbb{L}_{CE}}(a_{ct}, \tilde{a}_{ct}) + \lambda \mathop{\mathbb{L}_{MAE}}(a_{cs}, \tilde{a}_{cs}) \\
        \text{where } &\tilde{a}_{ct} =  C_T(F(x_v)), \quad \tilde{a}_{cs} = C_S( F(x_v), a_{h})
    \end{split}
    \label{eq:AC}
\end{equation}
and $x_v$ denotes the clothed human mask image, $a_{ct}$, $a_{cs}$, and $a_h$ denotes the cloth type, the cloth sizes in the cloth attribute, and the human attribute, respectively, $F$, $C_T$, $C_S$ are the feature extractor, cloth type classifier, and cloth size estimator, respectively, and $\mathop{\mathbb{L}_{CE}}$ and $\mathop{\mathbb{L}_{MAE}}$ denote the standard cross-entropy (CE) and mean absolute error (MAE) losses.

In the second stage of the proposed method, the U-Net-based AG-VTON is trained using the cloth auto-encoder and attribute classifier trained in the previous step. The virtual try-on network takes as input a silhouette image of the user, a target cloth mask image, and cloth and human attributes. An MLP-based feature extractor transforms the attribute vector, which is then reshaped and concatenated with the AG-VTON encoder's output and the feature vector extracted from the cloth auto-encoder. This controls the cloth area on the target human body. The synthesized virtual try-on mask is then fed into the pretrained attribute classifier to classify the cloth type and estimate cloth sizes. Thus, the virtual try-on network is trained to synthesize the clothed human mask to minimize the loss between the output of the network and ground truth, and the loss between the input cloth attributes and the cloth attributes estimated by the attribute classifier. Finally, the objective functions of the proposed method are defined as follows:
\begin{equation}
    \resizebox{0.91\columnwidth}{!}{
    $
    \begin{aligned}
        &\mathbb{L}_{VT} = \alpha \mathop{\mathbb{L}_{DICE}}(x_v, \tilde{x}_v) + \beta \mathop{\mathbb{L}_{CE}}(a_{ct}, \tilde{a}_{ct}) + \gamma \mathop{\mathbb{L}_{MAE}} (a_{cs}, \tilde{a}_{cs})\\
        &\text{where}~\mathop{\mathbb{L}_{Dice}}(p,q)= 1 - 2 \frac{\sum_i p^i q^i}{\sum_i p^i \sum_i q^i}, \quad \tilde{x}_v = G(x_h, x_c, a_c, a_h)
    \end{aligned}
    $
    }
    \label{eq:VTON}
\end{equation}
and $G$ is the AG-VTON, $x_h$ and $x_c$ denote the input user mask and the cloth mask, respectively, $a_c$ and $a_h$ are the cloth attribute and human attribute, respectively, and $\mathop{\mathbb{L}_{DICE}}$ denotes the dice loss, which is adopted for complex and highly imbalanced datasets to distinct properties of foreground and background.

\subsection{Style Transfer}
To generate realistic images of clothed human, we utilize a style transfer network since our proposed networks generate masks rather than texturized cloth.
We use a modified version of SieveNet \cite{jandial2020sievenet} for converting our cloth mask into textured cloth.
We replace the conditional segmentation mask generation module with our own modified version of the segmented mask, which is generated by superimposing the cloth mask from our model over the segmented mask of the whole body of the person generated using self-correction human parsing \cite{li2020self}. 
The modified segmented mask is used with a coarse-to-fine warping module and a segmentation-assisted texture translation module to generate the person wearing the cloth.

\section{Experimental Results}
\label{sec:experiments}
\begin{figure}[!t]
    \centering
    \includegraphics[width=\columnwidth]{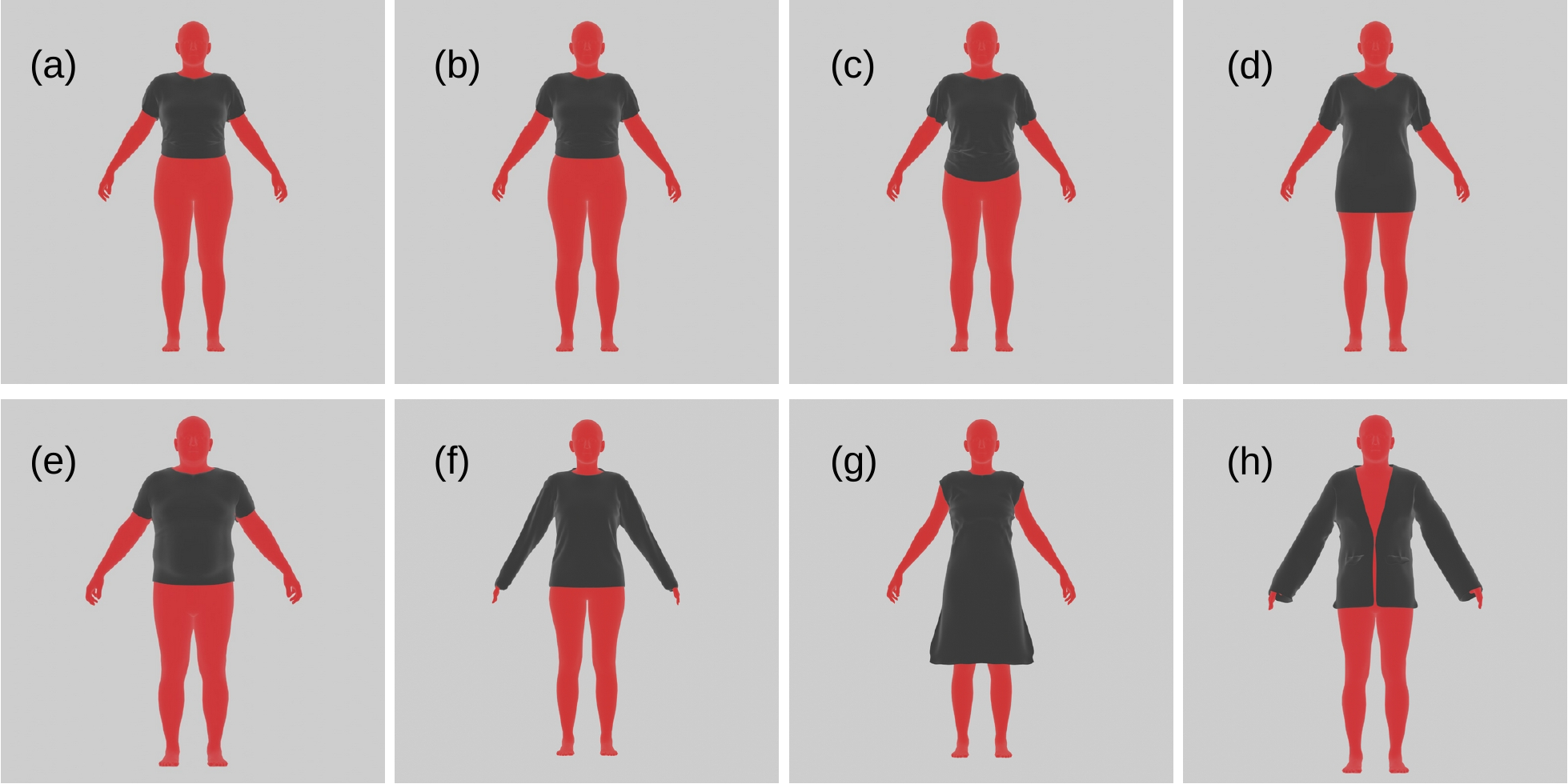}
    \vspace{-7mm}
    \caption{Example images of four size variations for the same t-shirt: (a) S, (b) M, (c) L, and (d) XL.
    Examples of four cloth types: (e) t-shirt, (f) long sleeve, (g) dress, and (h) blazer.}
    \vspace{-3mm}
    \label{fig:cloth_types}
\end{figure}

\textbf{Dataset:}
The cloth dataset used in this work consists of four different types of clothes: t-shirt, long-sleeve, dress, and blazer, as illustrated in \cref{fig:cloth_types}. For each type, we predefined the minimum and maximum size values. Random sizes were generated within these ranges, resulting in a diverse dataset containing various types and sizes of clothes. As depicted in \cref{fig:cloth_types}, the dataset contains clothes with different sizes, and the draping shape of each cloth varies depending on its size. 
Specifically, we generated 8,412 images of clothed females for t-shirts, long-sleeved shirts, and dresses, and 6,792 images of clothed males for t-shirts, long-sleeved shirts, and blazers.
In total, our proposed network was trained with 45,612 images.

\textbf{Implementation Details:}
For the evaluation of the virtual try-on mask image synthesis, we used F1, IoU scores, and average length error in pixel. To compare the proposed method with the state-of-the-art methods quantitatively, we use Frechet Inception Distances (FID) \cite{heusel2017gans} and Learned Perceptual Image Patch Similarities (LPIPS) \cite{zhang2018unreasonable}. The proposed networks were trained for 300 epochs with a batch size of 24, using Adam optimizer with a 0.001 learning rate. We implemented the proposed networks using PyTorch in NVIDIA GeForce RTX 3090.

\begin{table}[!t]
    \centering
    \caption{Quantitative evaluation of the AG-VTON and ablation result of attribute classifier}
    \vspace{-3mm}
    \resizebox{\columnwidth}{!}{
    \begin{tabular}{ | m{4em} | c | c | c | } 
          \hline
            & F1 $\uparrow$ & IoU $\uparrow$ & Average length error (pixel) $\downarrow$  \\ 
          \hline
          w/ AC & 0.9436 & 0.8933 & 11.7589 \\ 
          \hline
          w/o AC & 0.9409 & 0.8886 & 11.8511 \\ 
          \hline
    \end{tabular}}
    \vspace{-5mm}
    \label{tab:ac}
\end{table}


\textbf{Results on Synthetic Dataset:}
We evaluated the performance of the proposed method on a  data that is not used for training process. Our method achieved high F1 and IoU scores, as reported in \cref{tab:ac}. We also conducted ablation experiments to investigate the contribution of the attribute classifier. We found that training the network with the attribute classifier led to higher F1 and IoU scores, indicating improved accuracy in virtual try-on mask image synthesis. Additionally, the attribute classifier helped to predict accurate cloth sizes, as reflected in the lower average length-pixel difference. Specifically, we measured the number of pixels in the cloth length and observed that the attribute classifier effectively controlled the cloth length.

\begin{figure}[!t]
    \centering
    \includegraphics[width=\columnwidth]{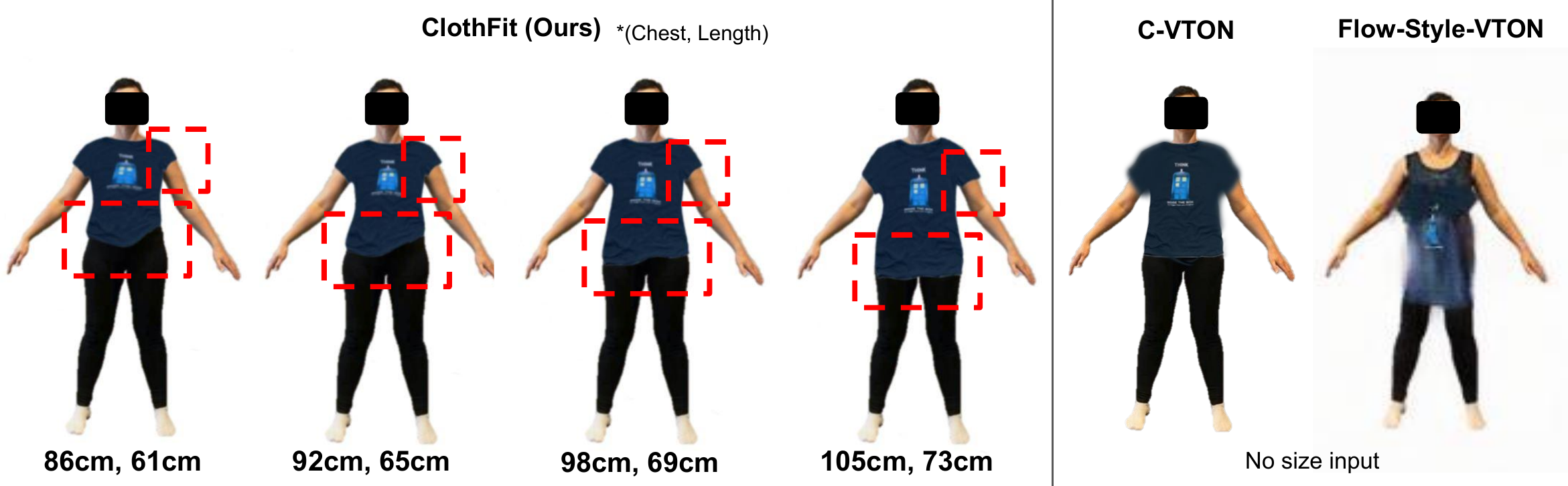}
    \includegraphics[width=\columnwidth]{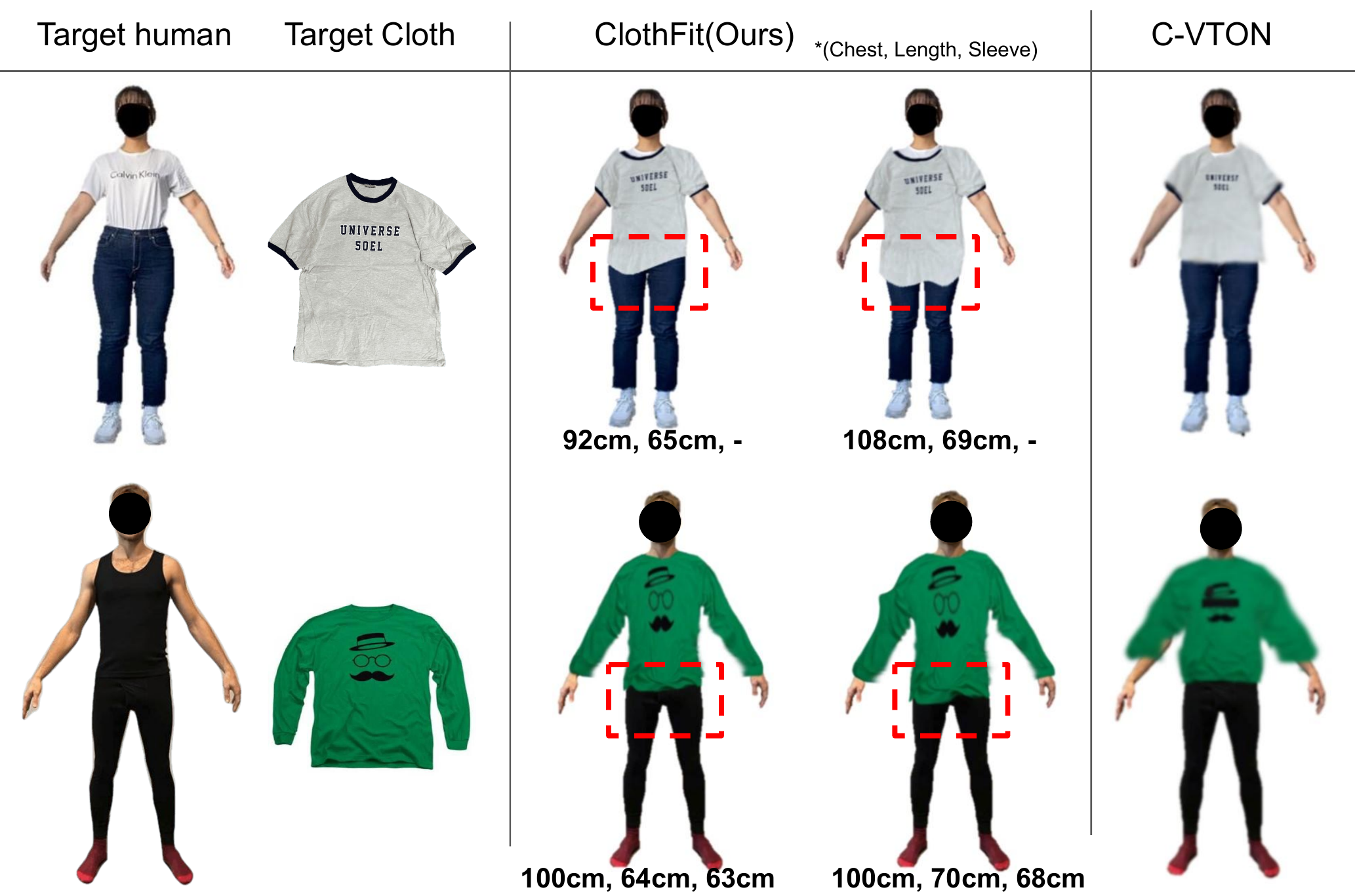}
    \vspace{-7mm}
    \caption{Result on real dataset: The same human wearing the same cloth with different sizes using our method and comparison result with state-of-the-art on the same dataset.}
    \vspace{-3mm}
    \label{fig:various_result}
\end{figure}

\textbf{Results on Real Dataset:}
\cref{fig:various_result} presents the qualitative evaluation with a real cloth and human image along with inference from other state-of-the-art models, such as C-VTON \cite{fele2022c} and Flow-Style-VTON \cite{he2022fs_vton}.
Current 2D virtual try-on methods follow a two-step process of image wrapping(estimation of target cloth shape) followed by texture transfer (estimation of target cloth texture).
To compare our model with previous state-of-the-art (SOTA) models, we replaced the first step of SieveNet with AG-VTON output and computed the  FID and LPIPS score that measures the similarity of texture alignment between the input and synthesized image.
As shown in \cref{tab:comparison}, our proposed method generates outputs of comparable quality to current SOTA methods and even achieves a lower LPIPS score than Flow-Style-VTON, although our model is not explicitly trained for texture translation.

\begin{table}[!t]
    \centering
    \caption{Quantitative comparison with the state-of-the-art method.}
    \vspace{-3mm}
    \resizebox{\columnwidth}{!}{
    \begin{tabular}{ | l | c | c | c |} 
          \hline
            & C-VTON \cite{fele2022c} & Flow-Style-VTON \cite{he2022fs_vton} & Clothfit (Ours) \\ 
          \hline
          FID $\downarrow$ & 19.54 & 33.87 & 40.76 \\ 
          \hline
          LPIPS $\downarrow$	 & 0.107 & 0.210 & 0.128 \\ 
          \hline
    \end{tabular}}
    \vspace{-5mm}
    \label{tab:comparison}
\end{table}

\section{Conclusion}
\label{sec:conclusion}
\vspace{-2mm}
In this work, we proposed a novel virtual try-on system, ClothFit, that generates photorealistic images of a person wearing a garment based on the actual size of the garment and human attributes. The system consists of a cloth auto-encoder, an attribute classifier, and a U-Net-based AG-VTON. We trained the proposed networks on the dataset generated using 3D physics simulation, Blender. The experimental results showed that the proposed method synthesized the virtual try-on images better than state-of-the-art models and could generate the virtual try-on images along with the actual human and cloth attributes. Our proposed system has the potential to be used in various industries such as e-commerce and fashion design, where the virtual try-on system can assist customers to try on garments virtually before purchasing, saving time and resources.

\bibliographystyle{IEEEbib}
\bibliography{refs}

\end{document}